\documentclass[letterpaper, 10 pt, conference]{ieeeconf}  

\IEEEoverridecommandlockouts                              

\usepackage{textcomp}
\usepackage{amsmath}
\usepackage{algorithm}
\usepackage{algpseudocode}
\usepackage{float}
\usepackage{stfloats}
\usepackage{tikz}
\usetikzlibrary{shapes.geometric, arrows, positioning}

\tikzstyle{main} = [rectangle, rounded corners, draw=black, fill=blue!20, text centered, text width=6cm, minimum height=1cm]
\tikzstyle{sub} = [rectangle, draw=black, fill=green!20, text centered, text width=5cm, minimum height=1cm]
\tikzstyle{arrow} = [thick,->,>=stealth]

\usepackage{cite}

\makeatletter
\let\NAT@parse\undefined
\makeatother
\usepackage[colorlinks,linkcolor=black,anchorcolor=blue,citecolor=blue]{hyperref}

\def\BibTeX{{\rm B\kern-.05em{\sc i\kern-.025em b}\kern-.08em
    T\kern-.1667em\lower.7ex\hbox{E}\kern-.125emX}}

\usepackage{graphics}
\usepackage{epsfig}
\usepackage{mathptmx} 
\usepackage{times}
\usepackage{amssymb}
\usepackage{ulem}
\usepackage{multirow}
\usepackage{threeparttable}
\usepackage{cleveref}
\crefname{table}{Table}{Tables}
\crefname{figure}{Fig.}{Figs.}
\crefname{section}{Section}{Sections}

\usepackage{url} 
\usepackage{booktabs}
\usepackage{graphicx} 
\usepackage{tcolorbox}
\usepackage{svg}
\usepackage{hyperref}
\usepackage[acronym]{glossaries}
\newacronym{mlp}{MLP}{Multilayer Perceptron}
\newacronym{svm}{SVM}{Support Vector Machine}
\newacronym{knn}{KNN}{K-Nearest Neighbors}
\newacronym{gb}{GB}{Gradient Boosting}
\newacronym{rf}{RF}{Random Forest}
\newacronym{fpr}{FPR}{False positive rate}
\newacronym{tpr}{TPR}{True positive rate}
\newacronym{fnr}{FNR}{False negative rate}
\newacronym{acc}{ACC}{Accuracy}
\newacronym{maxprob}{$\max(prob)$}{max probability}
\newacronym{minscore}{$\min(score)$}{min anomaly score}
\newacronym{prob}{$prob$}{probability}
\newacronym{score}{$score$}{anomaly score}
\newacronym{rl}{$RL$}{reinforcement learning}
\newacronym{smd}{SMD}{Scaled Manhattan Distanced Detector}
\newacronym{ocsvm}{OCSVM}{One-Class SVM}
\newacronym{km}{KM}{K-Means}
\newacronym{oc}{OC}{Outlier Counting Detector}
\newacronym{cp}{CP}{complex password}
\newacronym{np}{NP}{numeric password}
\newacronym{tp}{TP}{text-based password}
\newacronym{id}{ID}{user identification}
\newacronym{auth}{Auth}{user authentication}
\newacronym{nerf}{NeRF}{Neural Radiance Fields}
\newacronym{rlhf}{RLHF}{reinforcement learning from human feedback}
\newacronym{3d}{3D}{three-dimensional}
\newacronym{nbv}{NBV}{next best view}
\newacronym{psnr}{PSNR}{Peak Signal-to-Noise Ratio}
\newacronym{ssim}{SSIM}{Structural Similarity Index}
\newacronym{lpips}{LPIPS}{Learned Perceptual Image Patch Similarity}
\newacronym{ppo}{PPO}{Proximal Policy Optimization}
\newacronym{drl}{DRL}{Deep Reinforcement Learning}
\newacronym{ftp}{FTP}{File Transfer Protocol} 
\newacronym{dof}{DOF}{degrees of freedom}
\makeglossaries

\title{\LARGE \bf
Preference-Driven Active 3D Scene Representation for Robotic Inspection in Nuclear Decommissioning
}

\author{
    Zhen Meng$^{1}$, Kan Chen$^{1}$, Xiangmin Xu$^{1}$, 
    Erwin Jose Lopez Pulgarin$^{2}$, Emma Li$^{1}$, 
    \\Philip Guodong Zhao$^{3}$, and David Flynn$^{4}$
\thanks{$^{1}$Zhen Meng, Kan Chen, Xiangmin Xu and Emma Li are with the School of Computing Science, University of Glasgow, G12 8RZ, Glasgow, UK.
        {\tt\small \{zhen.meng, liying.li\}@glasgow.ac.uk, \{k.chen.1, x.xu.1\}@research.gla.ac.uk.}}%
\thanks{$^{2}$Erwin Jose Lopez Pulgarin is with the Department of Engineering, University of Manchester, M13 9PL, Manchester, UK.
        {\tt\small erwin.lopezpulgarin@manchester.ac.uk.}}%
\thanks{$^{3}$Philip Zhao is with the Department of Computer Science, University of Manchester, M13 9PL, Manchester, UK.
        {\tt\small philip.zhao@manchester.ac.uk.}}%
\thanks{$^{4}$David Flynn is with James Watt School of Engineering, University of Glasgow, G12 8QQ, Glasgow, UK
        {\tt\small david.flynn@glasgow.ac.uk.}}%
}

\begin{document}

\maketitle
\thispagestyle{empty}
\pagestyle{empty}

\begin{abstract}
Active 3D scene representation is pivotal in modern robotics applications, including remote inspection, manipulation, and telepresence. Traditional methods primarily optimize geometric fidelity or rendering accuracy, but often overlook operator-specific objectives, such as safety-critical coverage or task-driven viewpoints. This limitation leads to suboptimal viewpoint selection, particularly in constrained environments such as nuclear decommissioning. To bridge this gap, we introduce a novel framework that integrates expert operator preferences into the active 3D scene representation pipeline. Specifically, we employ Reinforcement Learning from Human Feedback (RLHF) to guide robotic path planning, reshaping the reward function based on expert input. To capture operator-specific priorities, we conduct interactive choice experiments that evaluate user preferences in 3D scene representation. We validate our framework using a UR3e robotic arm for reactor tile inspection in a nuclear decommissioning scenario. Compared to baseline methods, our approach enhances scene representation while optimizing trajectory efficiency. The RLHF-based policy consistently outperforms random selection, prioritizing task-critical details. By unifying explicit 3D geometric modeling with implicit human-in-the-loop optimization, this work establishes a foundation for adaptive, safety-critical robotic perception systems, paving the way for enhanced automation in nuclear decommissioning, remote maintenance, and other high-risk environments.


\end{abstract}
\section{Introduction}
Active \gls{3d} scene representation is a critical component of robotic vision, enabling a wide range of applications such as teleoperation~\cite{patil2024radiance}, navigation~\cite{chen2025splatnavsaferealtimerobot}, and manipulation~\cite{lu2024manigaussian}. In these scenarios, a robotic platform typically acquires visual data from multiple viewpoints and synthesizes a detailed representation of the scene. Most existing robotic viewpoint-planning methods rely on explicit geometric modelling or internal algorithmic parameters to guide decision-making~\cite{zhu20243dgaussiansplattingrobotics}. Although such approaches leverage metrics like geometric uncertainty or model residuals to inform the selection of \glspl{nbv}, they face several critical limitations.
\begin{itemize}
    \item Metric–User Mismatch: Quality measures based on novel-view rendering (e.g., \gls{psnr}, \gls{ssim}, \gls{lpips}) demand additional viewpoints for evaluation and often lack clear selection criteria~\cite{strong2024next}. Moreover, these metrics do not always align with real-world user requirements for 3D visualization, particularly when only specific regions or perspectives matter.
    \item Geometry–Visualization Gap: Optimizing \glspl{nbv} based on geometric accuracy or mesh completeness can theoretically enhance the fidelity of scene representations~\cite{ye2024pvp}. However, prioritizing highly accurate meshes does not always align with user expectations, particularly when the goal is to emphasize critical areas or distinctive structures rather than achieve full-scene completeness. Furthermore, geometric accuracy and visual rendering quality do not always correlate—an aesthetically pleasing rendering may not necessarily produce a clean or accurate mesh.~\cite{lee2022uncertainty}.
\end{itemize}
\begin{figure}[t]
    \centering
    \vspace{2mm}
    \includegraphics[width = 0.48\textwidth]{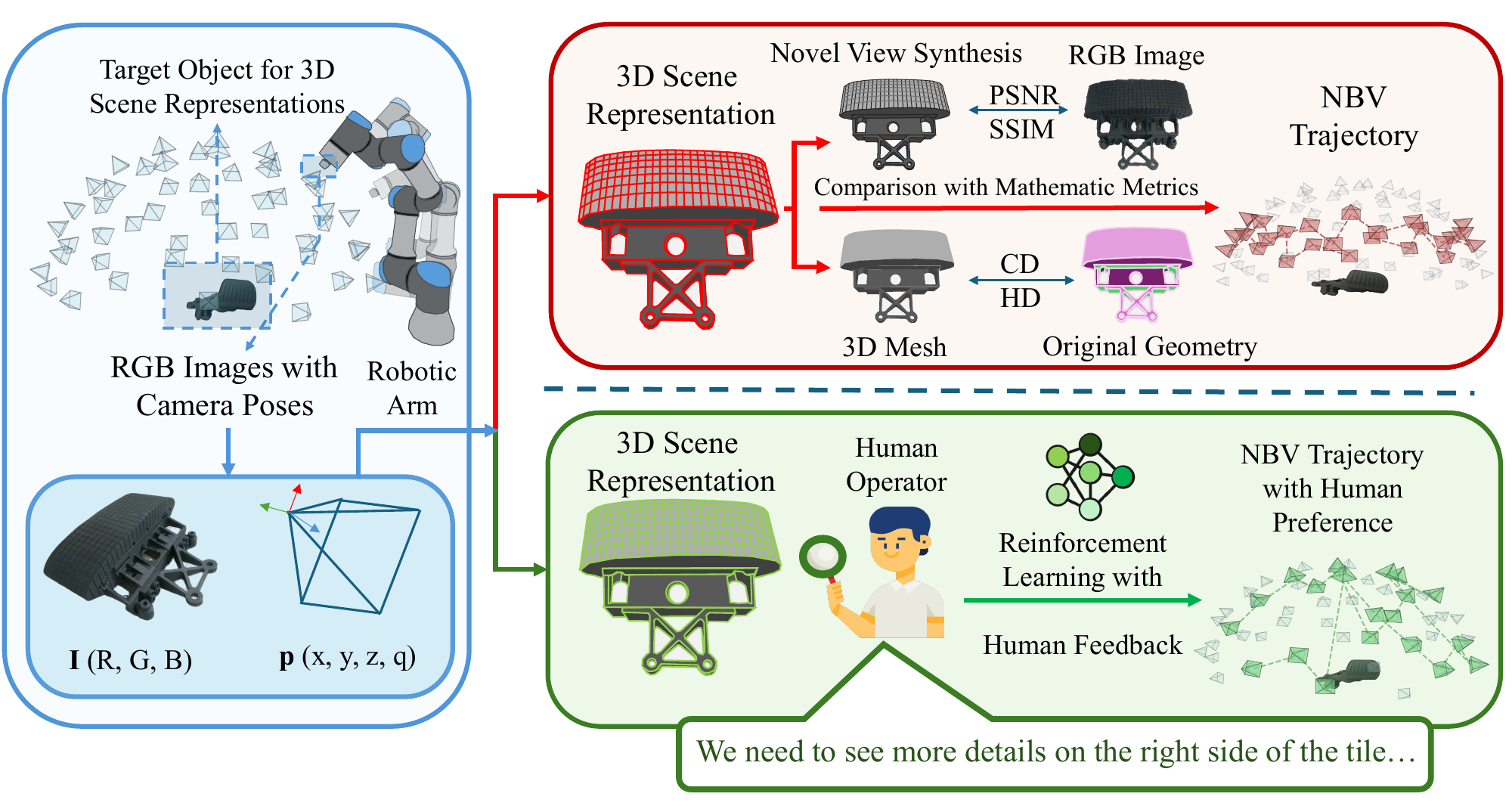}
    \caption{Motivation: Traditional methods rely on static metrics without considering task- and preference-based 3D representations.}
    \label{fig:motivation}
\end{figure}
In many industrial tasks involved with 3D scene representations, users require prioritized visualizations of specific regions, emphasizing user preferences over broad scene representation quality. Some task-specific objectives often prioritize user-centric visualization of critical regions rather than global scene representation accuracy~\cite{mitchell2023lessons}. For example, in teleoperated nuclear decommissioning, operators may need high-fidelity visualizations of a potentially hazardous material before interacting with it, while less critical background details can be simplified~\cite{10161011}. Similarly, in remote maintenance of nuclear facilities, ensuring clear visibility of occluded or safety-critical components, such as valve connections or radiation hotspots, takes precedence over globally consistent scene reconstruction~\cite{pacheco2021multiple}. Thus, existing metrics often do not match human perception of visual quality, highlighting a need for methods directly optimizing user experience. Even worse, in constrained environments such as the nuclear decommissioning process, where robot mobility and the data acquisition range are severely limited, full geometric information acquisition is not practical~\cite{lopez2022assessing}.

While human-in-the-loop approaches have been explored in nuclear RAS (Robotic and Autonomous Systems) and RAI (Robotics and Artificial Intelligence)~\cite{baniqued2024multimodal}, their integration into active 3D scene representation for optimizing viewpoint selection remains underdeveloped. Existing methods primarily focus on predefined heuristics or geometric accuracy, lacking adaptability to operator-driven priorities in safety-critical environments. On the one hand, users often prioritize specific regions or viewpoints of interest; on the other, standard metrics are unable to capture subjective requirements. While \gls{rlhf} has shown promise in natural language processing and 2D computer vision, its potential in robotics and 3D scene representations has not been fully realized~\cite{christiano2017deep}.

In this paper, we propose an \gls{rlhf}-based framework for active 3D scene representations that incorporates human preferences into viewpoint planning. As shown in Fig.~\ref{fig:motivation}, instead of relying on model-specific internal variables, our method draws on user evaluations of rendered 3D models to inform the policy-learning process. We demonstrate the effectiveness of this user-centric priority in a nuclear decommissioning scenario. The primary contributions of this work can be summarized as follows: 1) User-Centric \gls{nbv}: We introduce a \gls{drl}  strategy that explicitly accounts for the 3D visualization preference of the expert operator, thus eliminating reliance on fixed geometry- or rendering-based scores. 2) Scalable Policy Learning: Our approach integrates seamlessly with the robotic platforms and diverse 3D scene representations pipelines by avoiding dependence on algorithm-specific internal cues. 3) Empirical Validation: We evaluate our method on a real-world robotic testbed, showing that operator-guided policies significantly improve localized scene representations quality.

The rest of this paper is organized as follows. Section II reviews related studies on active \gls{3d} scene representations and user-centric optimization. Section III outlines the proposed \gls{rlhf} formulation and system architecture. Section IV presents the experiment implementation. Section V presents the baseline algorithm and experimental results, and Section VI concludes the paper and suggests directions for future research.

\section{Related Work}
Active \gls{3d} scene representations and \gls{nbv} planning have evolved considerably over the past decade. Early approaches focused on hand-crafted heuristics, employing volumetric coverage or geometric uncertainty metrics to guide viewpoint selection~\cite{641868},~\cite{chen2011active}. While these methods are effective in controlled or static environments, their reliance on predefined action spaces limits adaptability in complex, unstructured scenarios such as cluttered industrial settings or nuclear decommissioning facilities.

Recent advances leverage neural implicit representations, exemplified by \gls{nerf}, to achieve more expressive and continuous scene modelling~\cite{mildenhall2021nerf},~\cite{park2019deepsdf}. Although these representations facilitate high-fidelity scene representations and flexible novel view synthesis, they commonly assume dense multi-view inputs and high computational resources, challenges that constrain real-time or resource-limited robotic applications. To address these issues, methods like NeU-NBV integrate uncertainty estimation into neural rendering, enabling NBV planning that selectively focuses on less certain regions while reducing data requirements~\cite{jin2023neu}. Similarly, GenNBV employs \gls{drl} with multi-source feature embeddings to enhance generalization in large-scale and diverse environments~\cite{Chen_2024_CVPR}.

A key trend is the explicit incorporation of uncertainty modelling to guide active exploration. Techniques such as FisherRF use information-theoretic measures to select viewpoints without relying on ground-truth data~\cite{jiang2025fisherrf}. These strategies adaptively allocate sensing resources to informative directions, improving resilience against sparse observations and dynamic conditions. Such uncertainty-driven frameworks have proven beneficial for tasks like \gls{nerf}-based localization, where identifying high-value viewpoints can substantially reduce ambiguity in robot pose estimation~\cite{chen2024leveraging}. In sparse-view scenarios, balancing data efficiency and scene representation quality is critical. Approaches like PVP-Recon iteratively select viewpoints based on warping consistency scores, maintaining scene representations fidelity under limited data budgets~\cite{ye2024pvp}. Further, methods like SparseNeuS and ReVoRF leverage geometric or semantic priors to mitigate overfitting, thereby ensuring stable performance with minimal input views~\cite{long2022sparseneus},~\cite{xu2024learning}. However, these approaches are often benchmarked by conventional metrics (e.g.,\gls{psnr}, \gls{ssim}), which do not fully capture user preferences or task-specific requirements.

A promising research direction involves integrating human feedback into the \gls{nbv} planning loop while leveraging recent advances in embodied intelligence and large-scale language models. For example, the author in~\cite{10.1145/3613905.3650982} argues that humans can leverage contextual understanding to supplement missing data during the scene representations process. Therefore, a human-robot interaction mechanism is introduced, enabling remote users to participate in and guide the robot in completing the scene representations task.
Meanwhile, embodiment enables active physical interaction (e.g., moving or manipulating objects) to reduce occlusions and refine uncertain areas~\cite{jiang2024roboexpactionconditionedscenegraph}. Large-scale pretrained language models further enrich these pipelines by providing high-level reasoning and dynamic task planning in complex scenes~\cite{firoozi2023foundation}. For example, the author in~\cite{qi2024air} introduces an AIR-Embodied model that integrates embodied AI with multi-modal large language models to enable high-level reasoning, interactive object manipulations, and closed-loop verification, improving scene representations efficiency and generalization across diverse environments. Such approaches illustrate a shift toward user-centric, task-adaptive strategies, where traditional objective metrics are augmented by subjective preferences and operational constraints.

Despite these advancements, critical challenges persist—namely, resource limitations, environmental complexity, and the inherent difficulty of encoding subjective human goals directly into robotic policies. These gaps highlight the need for closer integration of uncertainty modelling, neural radiance representations, and \gls{rlhf} paradigms, ultimately fostering \gls{nbv} planning frameworks that are simultaneously robust, efficient, and aligned with the nuanced requirements of real-world robotic applications.

\section{Method}
\begin{figure*}[t]
    \centering
    \vspace{2mm}
    \includegraphics[width = 0.88\textwidth]{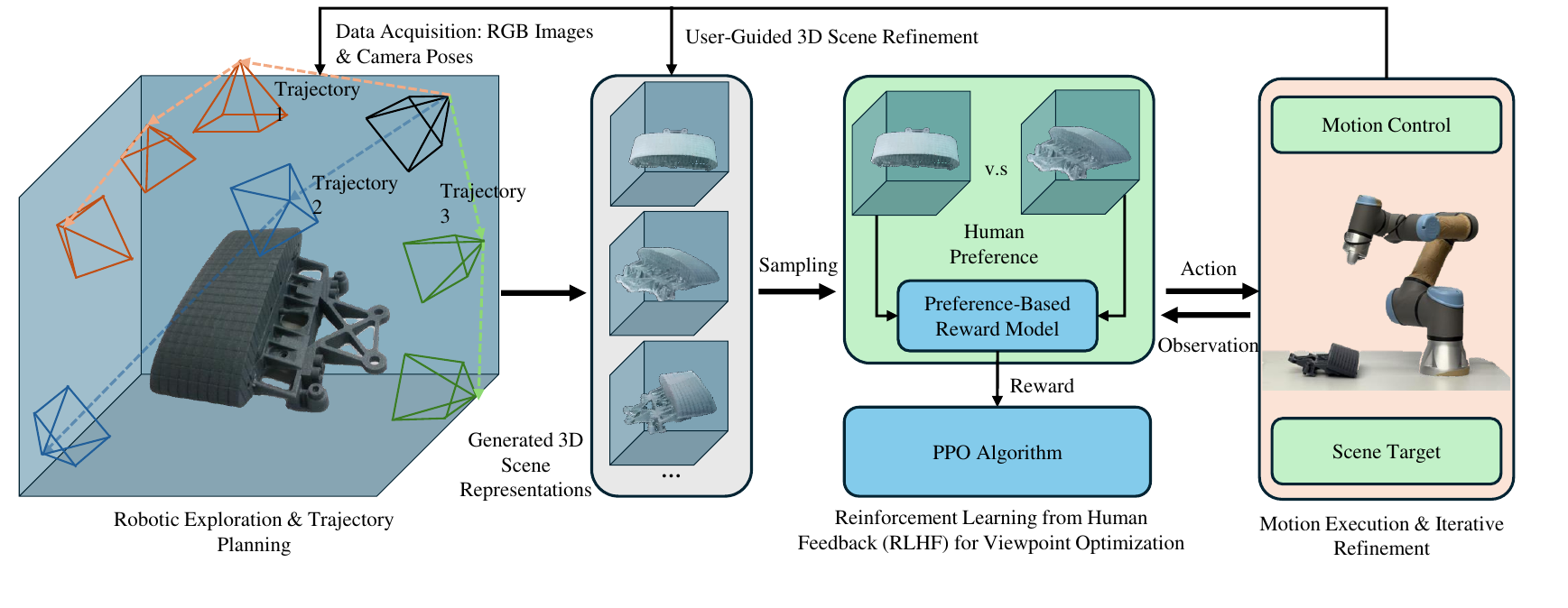}
    \caption{Overview of the proposed framework. The \gls{rlhf} pipeline consists of five key stages: (1) Robotic exploration and trajectory planning, where the robotic system collects observations; (2) Expert operator preference evaluation, where operators select preferred scene representations; (3) Learning a reward model based on collected human feedback; (4) Policy optimization using \gls{ppo} algorithms; and (5) Online training, where new data continuously refines the learned policy for improved viewpoint selection.}
    \label{fig:Framework}
\end{figure*}

\subsection*{Optimizing Active 3D Scene Representations via \gls{rlhf}}
Active 3D scene representations seek to optimize the actions of a robotic agent to generate high-quality 3D representations of objects or environments. This involves addressing two key challenges:
\begin{itemize}
    \item Exploration vs. Refinement: Balancing the discovery of new perspectives with revisiting regions to improve 3D scene representations accuracy.
    \item Quantifying Scene Representations Quality: Defining a reward function that captures 3D scene representations quality from a robotic perspective is inherently complex and subjective.
\end{itemize}
To address these challenges, as shown in Fig.~\ref{fig:Framework}, we propose leveraging \gls{rlhf} to integrate the preference of the expert operator into the optimization process, enabling the agent to learn a reward function aligned with subjective quality assessments.

\subsubsection*{\textbf{Step 1: Robotic Exploration and Trajectory Planning}}
At each time step $t$, the robot agent performs the following steps:

\begin{itemize} \item \textbf{Observation:} The agent captures an RGB image using its camera and extracts relevant features through the YOLO v11 model~\cite{khanam2024yolov11}. We selected YOLO v11 due to its real-time inference capability and strong feature extraction performance. We used a pre-trained YOLO v11 model without fine-tuning, as its generalization ability to common object features was sufficient for our scene representation task. The processed observation is represented as $o_t \in \left[-1,1\right]^{d_1 \times d_2 \times d_3}$, where $d_1$, $d_2$, and $d_3$ denote the dimensions of the extracted feature map.

\item \textbf{Action:} The agent selects an action $a_t$ from the discrete action space $\mathbb{A} = \{1,2,3,\dots,\mathcal{A}\}$, where $\mathcal{A}$ represents the total number of possible camera poses and corresponding image acquisitions.

\item \textbf{Update:} The robotic arm moves within the workspace according to the selected action. Upon reaching the new pose, a fresh observation $o_t$ is captured and stored in the dataset.
\end{itemize}

The interaction process generates a trajectory $\sigma = {(o_t, a_t)}_{t=1}^k$, which consists of a sequence of $k$ observation-action pairs. These trajectories serve as the foundation for subsequent reward evaluation and policy learning.

    

\subsubsection*{\textbf{Step 2:Evaluating the Preferences of Expert Operators}}
\begin{figure*}[t]
    \centering
    \vspace{2mm}
    \includegraphics[width = 1\textwidth]{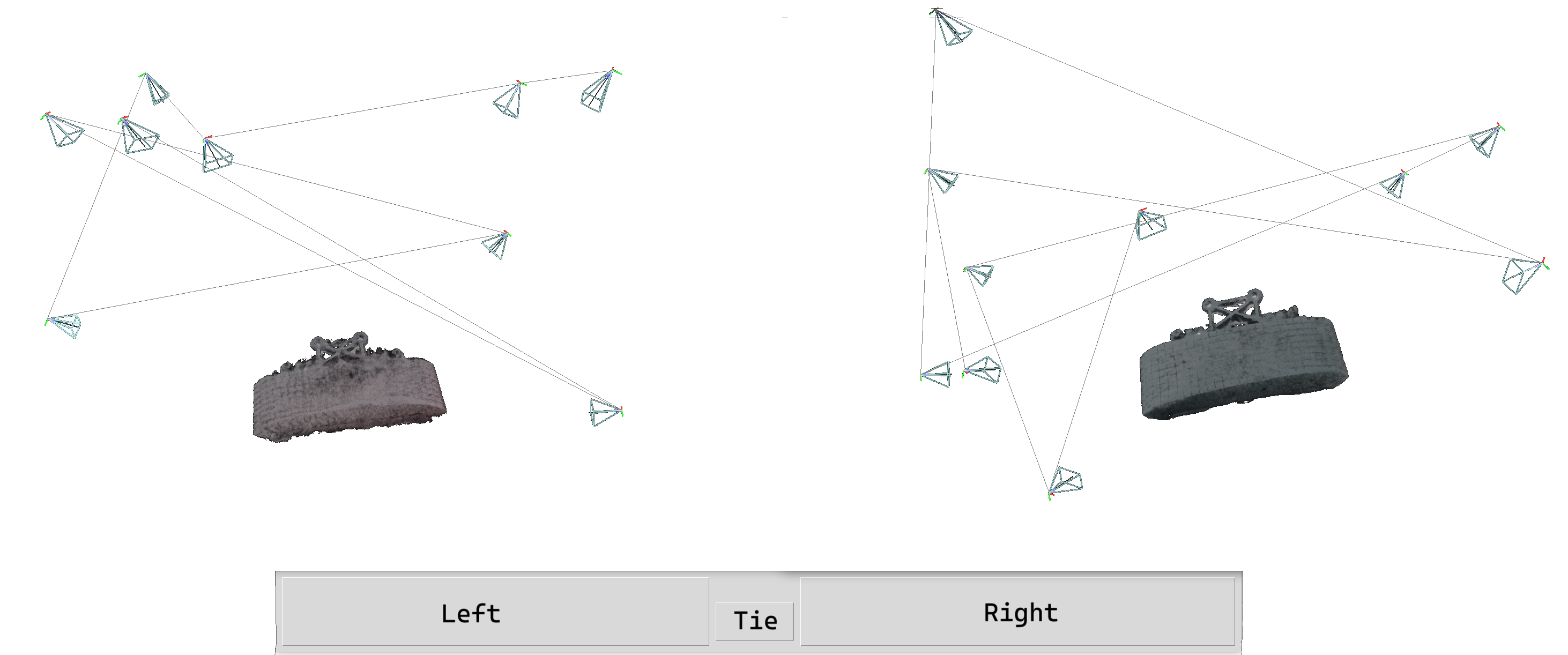}
    \caption{Illustrations of user interface for preference-based 3D scene selection and sequence representation. The interface allows users to compare and select preferred 3D scene representations, which are used to train a reward predictor for viewpoint optimization. Users can zoom, rotate, and inspect models for detailed evaluation. Notably, the illustrated line indicates the viewpoint selection order, not actual robotic motion. The real UR3e motion is planned using the Isaac Sim motion planner for smooth and optimized execution.}
    \label{fig: UI}
\end{figure*}

To ground the quality of scene representations in expert operator preference:
\begin{itemize}
    \item Model Comparison: As shown in Fig.~\ref{fig: UI}, the expert operator compares pairs of 3D scene representations $M_1$ and $M_2$, derived from trajectory segments $\sigma_1$ and $\sigma_2$, respectively. Since the focus is on task-relevant scenes or objects, we employed the Segment Anything Model (SAM) to segment the target regions of interest~\cite{kirillov2023segment}. To ensure accurate segmentation aligned with the expert operator’s comparison process, we fine-tuned SAM using images from the first-step observation, enabling more precise scene representation tailored to the task.
    \item Preference Indication: Evaluators provide a binary preference $\mu \in \{1, 2\}$, signifying which scene representations better captures the object or environment.
\end{itemize}

\subsubsection*{\textbf{Step 3: Reward Function Estimation}}
A reward function $\hat{r}(o_t, a_t)$ is trained to predict the expert operator preferences~\cite{christiano2017deep}:
\begin{itemize}
    \item Preference Modelling: Define the likelihood of the expert operator preference for trajectory $\sigma_1$ over $\sigma_2$ as:
    \[
    P[\sigma_1 \succ \sigma_2] = \frac{\exp \left(\sum_{t=1}^k \hat{r}(o_t^1, a_t^1)\right)}{
    \exp \left(\sum_{t=1}^k \hat{r}(o_t^1, a_t^1)\right) + \exp \left(\sum_{t=1}^k \hat{r}(o_t^2, a_t^2)\right)}.
    \]
    \item Loss Function: Train $\hat{r}$ by minimizing the cross-entropy loss:
    \begin{align}
    \text{Loss}(\hat{r}) = 
    & -\sum_{(\sigma_1, \sigma_2, \mu) \in D} \big[ 
    \mu(1) \log P[\sigma_1 \succ \sigma_2] \nonumber \\
    & + \mu(2) \log P[\sigma_2 \succ \sigma_1]
    \big],
    \end{align}
    where $D$ is the dataset of trajectory pairs and preferences.
\end{itemize}
\subsubsection*{\textbf{Step 4: Policy Optimization}}
Using the learned reward function $\hat{r}$, we optimize a policy $\pi$ to maximize cumulative rewards:
\[
R = \sum_{t=1}^T \hat{r}(o_t, a_t).
\]
In this study, we employ \gls{ppo} as the primary \gls{drl}  algorithm~\cite{ppo}. However, our approach is not restricted to \gls{ppo} and can be generalized to a wide range of \gls{drl} algorithms.

\subsubsection*{\textbf{Step 5: Online Training for Fine-tuning}}
After executing the actions dictated by the optimized policy, the robotic arm follows the planned trajectory to generate a 3D scene representation model. The newly generated representations can then be incorporated into the user preference selection process, enabling continuous refinement through online training. As the robotic arm moves, it captures images of regions with a higher probability of user interest, along with corresponding camera poses. These newly acquired images and poses can be fed back into the first step to generate additional 3D scene representations, further enriching the dataset. This process can be iteratively repeated, progressively optimizing the robotic motion trajectory and improving both viewpoint efficiency and scene representation quality over time.
\begin{figure}[t]
    \centering
    \vspace{2mm}
    \includegraphics[width = 0.49\textwidth]{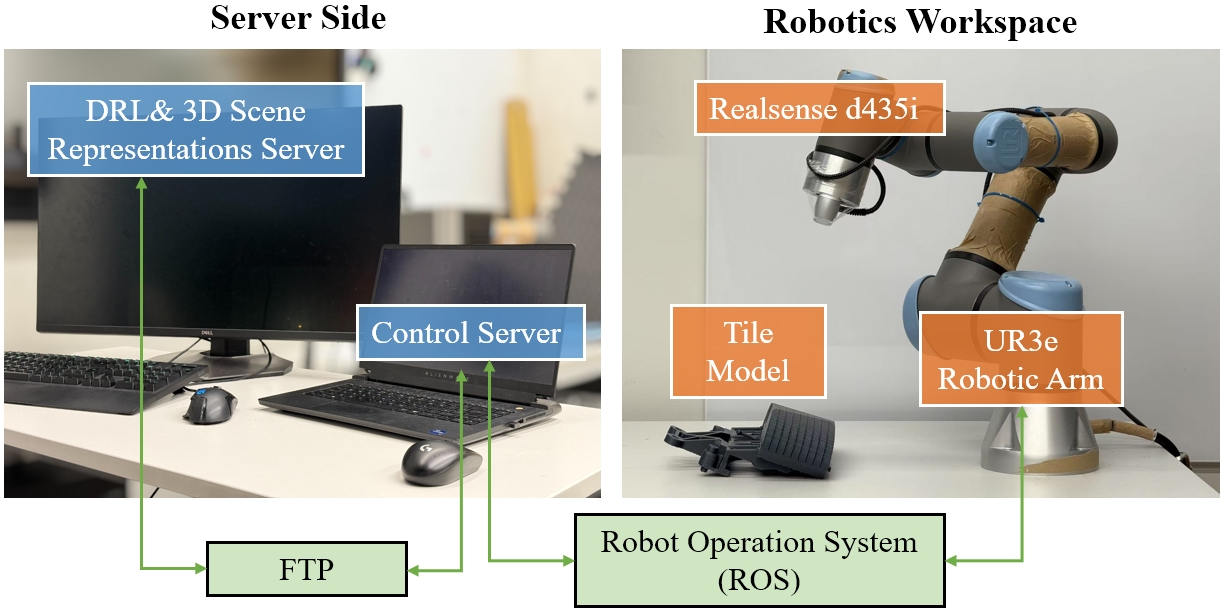}
    \caption{Experimental setup of the \gls{rlhf}-based 3D scene representation system.
    The setup consists of a UR3e robotic arm with an Intel RealSense D435i camera, controlled via a ROS-based framework. A control server handles motion execution, while a \gls{drl} server optimizes viewpoint selection based on human feedback. An \gls{ftp} ensures efficient data transfer, enabling real-time policy refinement for 3D scene representation. Our demo video is available at \url{https://youtu.be/mAAipFOotx8}.}
    \label{fig: testbed setup}
\end{figure}

\section{Experiment Setup}

\subsection{Robotic System Configuration}
To validate our \gls{rlhf}-based active 3D scene representations framework, as shown in Fig.~\ref{fig: testbed setup}, we employed a Universal Robots UR3e robotic manipulator~\cite{ur3e}, with an Intel RealSense d435i camera mounted on its end-effector~\cite{Intel}. Although this camera is capable of capturing depth data, only the RGB output is utilized, aligning with our three 3D scene representations baseline method, aligned with the baseline methods in Section V-A. The UR3e offers six \gls{dof} and achieves a repeatability of $\pm0.03\,\text{mm}$, ensuring precise and consistent positioning during viewpoint adjustments.

To accommodate real-time robot control and the computationally intensive 3D scene representations processes, our testbed relies on two high-performance servers:
\begin{itemize}
    \item \textbf{Robotic Arm Control Server:} 
    Executes real-time motion trajectories and processes control signals. It leverages rmpflow motion generation tools in the \textit{NVIDIA Isaac Sim}~\cite{sim} for trajectory optimization, and use \textit{MoveIt} Kinematics Plugin for smooth and accurate joint motion execution~\cite{ros_moveit}. The hardware setup includes an \textit{Intel Core i7-11700 CPU, 32\,GB RAM, and an NVIDIA RTX 3070 GPU}, enabling low-latency motion control and on-the-fly image acquisition.
    \item \textbf{\gls{drl} and 3D Scene Representations Server:} 
    Hosts the \gls{rlhf} training and the scene representations algorithm. An \textit{Intel Core i9-13900 CPU, 64\,GB RAM, and an NVIDIA RTX 4090 GPU} enables efficient policy optimization and real-time rendering. 
\end{itemize}
A Robot Operating System (ROS)-based framework coordinates communication among the robotic arm, RGB cameras, and control modules. The control server transmits motion commands and receives sensor feedback via ROS topics, ensuring synchronized execution.  The \gls{drl} and Scene Representations Server receives images captured by the camera and the end-effector's pose from the control server via the \gls{ftp}. It then sends the target pose of the end-effector back to the control server. The \gls{ftp} facilitates large data transfers between servers, particularly for scene reconstruction and \gls{drl}  updates. The real-world robotic workspace integrates a robotic arm, RGB cameras, and an operational tile-based environment, ensuring robust testing conditions.


\subsection{Experiment Implementations}
We focus on a critical task in the automation process of nuclear decommissioning: utilizing a robotic arm for high-precision 3D visualization and inspection of reactor tiles~\cite{chapman2019ukaea}. These tiles, typically composed of beryllium-coated Inconel and arranged in a slightly tilted manner to regulate plasma circulation within the reactor core, play a crucial role in maintaining structural integrity. Damage to these tiles can lead to plasma leakage and further degradation of internal reactor components, making accurate inspection essential for safe decommissioning operations~\cite{Austin-Morgan_2022}.

To facilitate this process, we employ high-precision, full-scale 3D-printed replicas of reactor tiles as test samples, ensuring a realistic and controlled evaluation environment. The designed pipeline systematically acquires RGB images and corresponding poses from multiple viewpoints, integrating the preference of the expert operator to refine the \gls{rlhf}-based policy for enhanced inspection accuracy. The details are as follows:


\begin{enumerate}
\item \textbf{Scene Initialization and Calibration:}
A structured environment is established within the UR3e workspace to simulate realistic 3D scene representation scenarios. Before data collection, hand-eye calibration is performed to accurately determine the spatial relationship between the camera and the robot's base frame~\cite{hand-eye}. This process includes calibrating the camera's intrinsic parameters to ensure precise image measurements, while also estimating the camera's pose relative to the robotic arm. By obtaining an accurate transformation between the camera and the robot base, the system can achieve high-precision 3D scene representations. In addition, this calibration ensures that the robotic arm can move to the required poses with precision, enabling reliable perception and interaction within the workspace.

\item \textbf{Preference of Expert Operator:} One experienced personnel collaborating with Remote Applications in Challenging Environments, UK Atomic Energy Authority (UKAEA)~\cite{ukaea}, who possess in-depth knowledge of nuclear decommissioning environments and 3D scene representations quality requirements, is selected as the expert operator. The expert is familiar with the objectives of scene reconstruction, including clarity, accuracy, and task-specific relevance. Although the experiment involves only a single subject, its reliability is enhanced through multiple repetitions and various controlled experiments. Furthermore, an in-depth discussion on the differences in personalized requirements among multiple users is beyond the scope of this study. Then, multiple 3D scene reconstructions, generated from different viewpoint trajectories, are presented to the expert operators for comparative evaluation. Users interact with the models through zooming and rotation to conduct a detailed assessment. Preferences are selected based on task-relevant criteria such as focus area, visual clarity, feature coverage, and geometric fidelity. These pairwise comparisons are aggregated to train a reward model that encapsulates human-centric quality judgments. Every 10 captures, a new 3D scene representations is generated, repeating for 400 instances under varying viewpoints and conditions. These reconstructions are then paired into 200 comparison groups for the expert operator's evaluation.


\item \textbf{Policy Optimization:} 
Using \gls{ppo}, the \gls{drl} agent updates its viewpoint-selection policy to maximize the reward function derived from the preference of the expert operator data. 
The training parameters for \gls{drl} can be found in TABLE I. In addition, the agent autonomously refines its viewpoint selection policy through an iterative learning process. Initially, it explores the environment using random viewpoint sampling to collect a diverse set of observations. As learning progresses, the policy is progressively optimized based on the expert operator preference, guiding the selection toward viewpoints that align with user preferences. By leveraging \gls{rlhf}, the system adapts to prioritize views that enhance critical scene representation quality while maintaining efficiency in data collection and robotic motion.

\end{enumerate}

\section{Results}
\label{sec:Results}
\subsection{Baseline 3D Representation Methods}
We evaluate three 3D scene representation approaches to establish broader performance baselines,
\begin{table}[t]\label{parameters}
\renewcommand{\arraystretch}{1}
\centering
\caption{RL Parameters for Performance Evaluation}
\begin{threeparttable} 

\begin{tabular}{|l|l|}
\hline 
\bf{Parameters} & \bf{Values}\\
\hline
photos taken per reconstruction & 10\\
\hline
learning rate & $1e-5$\\
\hline
batch size & {32} \\
\hline
n-step return & {256} \\
\hline
discount factor & $0.99$\\
\hline
value function coefficient & $0.99$\\
\hline
clip range  & 0.2\\
\hline
\end{tabular}

\end{threeparttable}
\label{tab:rl_param}
\end{table}

\begin{itemize}
    \item \textbf{Instant-NGP~\cite{mueller2022instant}:} 
    A \textit{\gls{nerf}-based} method that uses multi-resolution hash encoding for rapid and high-fidelity scene representations. We adopt it as the main reference point for implicitly modelling radiance fields directly from RGB images.
    \item \textbf{3DGS~\cite{kerbl20233d}:} 
    An advanced volumetric approach designed to capture complex scene geometry. It leverages a grid-based structure and tailored optimization strategies to handle varying scene complexities.
    \item \textbf{PGSR~\cite{Chen_2024}:} 
    A geometry-optimized scene representations pipeline that emphasizes precise surface modelling. By focusing on explicit geometric cues, PGSR aims to achieve accurate mesh representations, particularly in scenarios demanding fine-grained detail.
\end{itemize}
 All three methods are integrated with identical data-capture protocols, enabling a consistent evaluation of their strengths, weaknesses, and overall scene representation quality when operating on RGB images as they encompass implicit, volumetric, and explicit modelling paradigms. By covering these distinct methodologies, our selection provides a diverse and representative evaluation framework, demonstrating that our \gls{rlhf}-based approach can effectively leverage existing well-established benchmarks for broad applicability.

\subsection{Evaluation Metrics}
Evaluation of our proposed framework combines both objective quality assessments and subjective alignment with the preference of the expert operator:

\begin{itemize}
    \item \textbf{Scene Representations Fidelity:}
    \begin{itemize}
        \item \textit{\gls{psnr}:} Quantifies the alignment between rendered novel views and reference images.
        \item \textit{\gls{ssim}:} Measures perceptual consistency in local intensity and texture patterns.
        \item \textit{\gls{lpips}:} Uses deep network embeddings to gauge perceptual similarity between synthetic and real images.
    \end{itemize}

    \item \textbf{Trajectory Efficiency:}
    \begin{itemize}
        \item \textit{Robot path length:} The distance travelled by the end-effector, reflecting mechanical and time efficiency in the exploration.        
    \end{itemize}
    
    \item \textbf{Human Preference Alignment:}
\begin{itemize}
    \item \textit{Subjective Visual Assessment:} Evaluators compare 3D scene renderings from the optimized policy and a random strategy, assessing clarity, completeness, and detail.  
    \item \textit{Comparison with Random Strategy:} Evaluators reviewed paired renderings and indicated which provided better scene understanding.
\end{itemize}
\end{itemize}

\begin{table}
\centering
\caption{Comparative Analysis of Proposed algorithm Performance Across Different 3D Representation Methods}
\label{table_length}
\resizebox{0.95\linewidth}{!}{
  \begin{tabular}{c|c|c|c|c|c|c}
    \hline    
     & \multicolumn{3}{|c}{\textbf{Ramdom Policy}} & \multicolumn{3}{|c}{\textbf{Proposed Algorithm}}\\
    \hline    
    \textbf{Approaches} & Instant-ngp & 3DGS & PGSR & Instant-ngp & 3DGS & PGSR\\
    \hline
    Average Path Length (m) &  3.28 & 3.28 & 3.28 &  2.25 & 2.49 & 2.43  \\
    \hline
    PSNR &  36.27 & 38.32  & N/A & 39.33 & 39.54 & N/A  \\
    \hline
    SSIM &  0.972 & 0.9341 & N/A & 0.985 & 0.979 & N/A  \\
    \hline
    LPIPS & 0.0134  & 0.0201 & N/A & 0.0109 & 0.0126 & N/A  \\
    \hline
  \end{tabular}
}
\end{table}

By combining these objective and human-informed metrics, we demonstrate that our \gls{rlhf}-based method not only achieves high-fidelity 3D scene representations from RGB images alone, but also aligns closely with human judgments of scene quality.

\begin{figure}[t]
    \centering
    \vspace{2mm}
    \includegraphics[width = 0.48\textwidth]{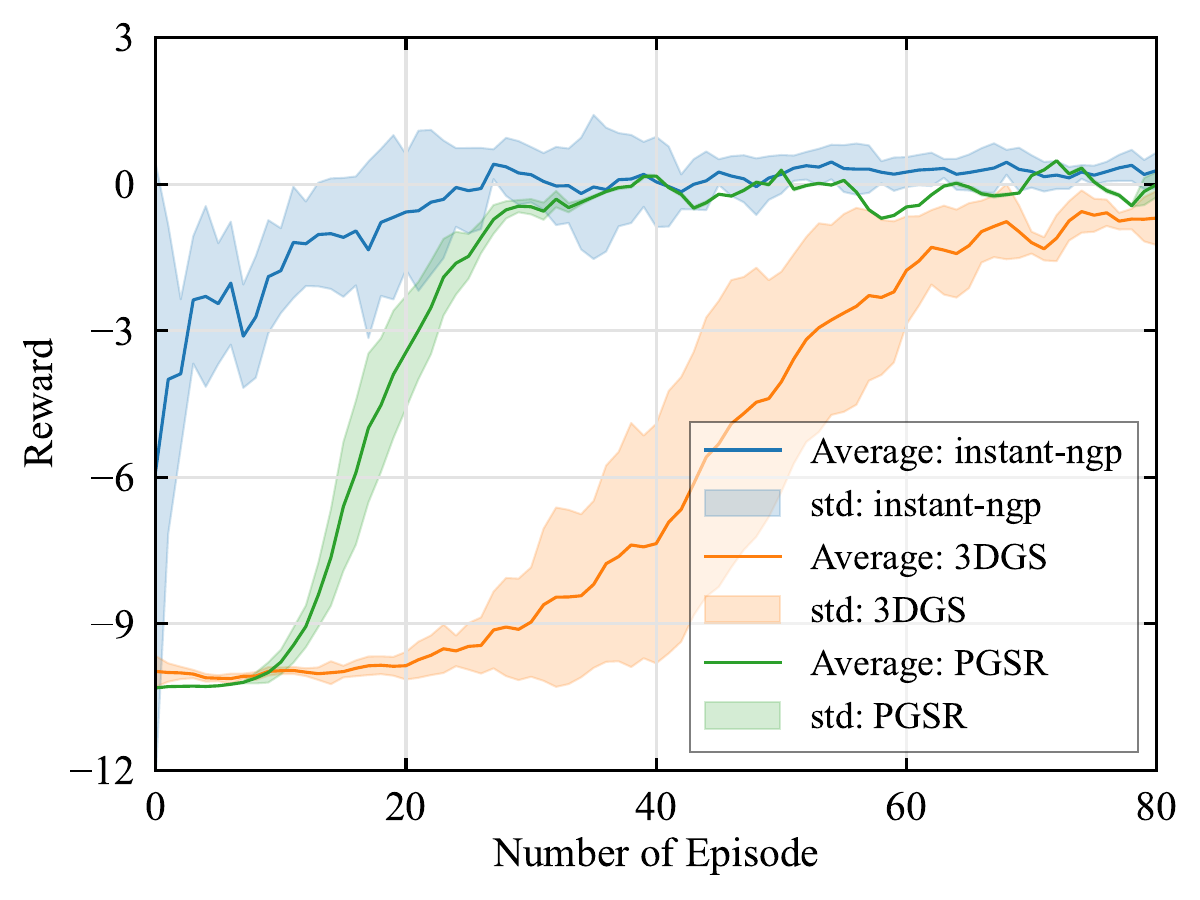}
    \caption{Convergence performance of our proposed framework across different 3D scene representation methods.}
    \label{fig: Reward}
\end{figure}

\begin{figure}[t]
    \centering
    \vspace{2mm}
    \includegraphics[width = 0.48\textwidth]{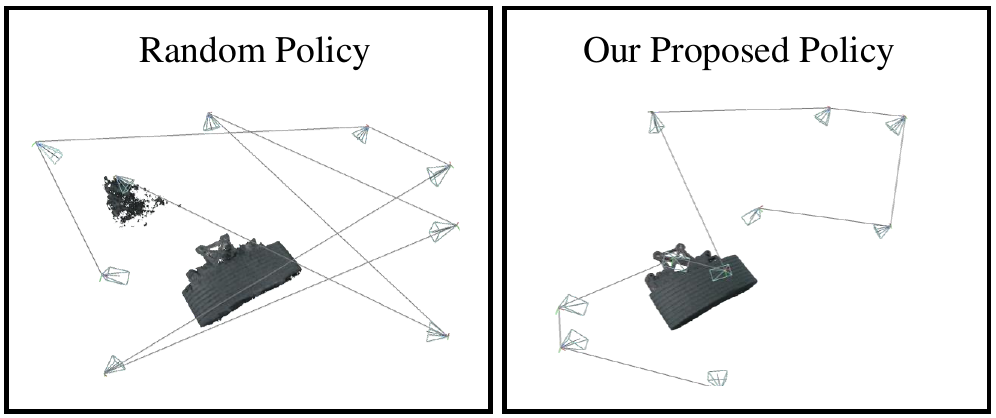}
    \caption{Trajectory efficiency comparison between our proposed algorithm and random policy.}
    \label{fig: length}
\end{figure}

\subsection{Convergence of \gls{rlhf} Across Different 3D Representation Methods}
Our proposed \gls{rlhf} framework demonstrates strong convergence behavior when trained on three different 3D representation methods: Instant-NGP, 3DGS, and PGSR. As shown in Fig.~\ref{fig: Reward}, our proposed framework achieves stable policy updates efficiently across all three methods, indicating its robustness and adaptability to different scene representations. The convergence results validate the effectiveness of incorporating human preference feedback, which accelerates learning and improves overall scene representation quality. To ensure statistical reliability, each algorithm was repeated five times, and the standard deviation (std) was computed to verify the robustness of the proposed approach. This demonstrates that our proposed framework can effectively guides the robotic system to capture viewpoints that maximize visual clarity and align with user-centric representation.


\subsection{Comparative Performance Evaluation}

To quantify the effectiveness of \gls{rlhf}, we evaluate multiple metrics, including \gls{psnr}, \gls{ssim}, and \gls{lpips}, across different representation methods. TABLE~\ref{table_length} presents a comparative analysis where our proposed algorithm demonstrates. The metrics are obtained from an average of comparisons between the ground truth image from the training dataset and the rendered novel view at the same camera pose. The novel views are specifically sampled near regions of interest identified by the expert operator, ensuring that the evaluation focuses on areas critical for operational and inspection tasks. Notably, \gls{rlhf} achieves higher \gls{ssim} scores, indicating better perceptual similarity, and lower \gls{lpips} values, reflecting improved feature alignment with human perception, particularly in regions of highest operational and inspection significance. Additionally, trajectory efficiency analysis reveals that \gls{rlhf} requires fewer robotic movements while maintaining optimal scene reconstruction quality, demonstrating its effectiveness in reducing operational cost and computational overhead. As shown in Fig.~\ref{fig: length}, we sampled trained instant-ngp robotic end-effector trajectories and compared them with random policy. The results indicate that our proposed algorithm exhibits a more efficient movement sequence, leading to smoother and more structured trajectory execution. We also summarized the average path length in the TABLE~\ref{table_length} by computing multiple trajectory datasets.

\subsection{Local Rendering Quality vs. Random Strategies}
\begin{figure}[t]
    \centering
    \vspace{2mm}
    \includegraphics[width = 0.46\textwidth]{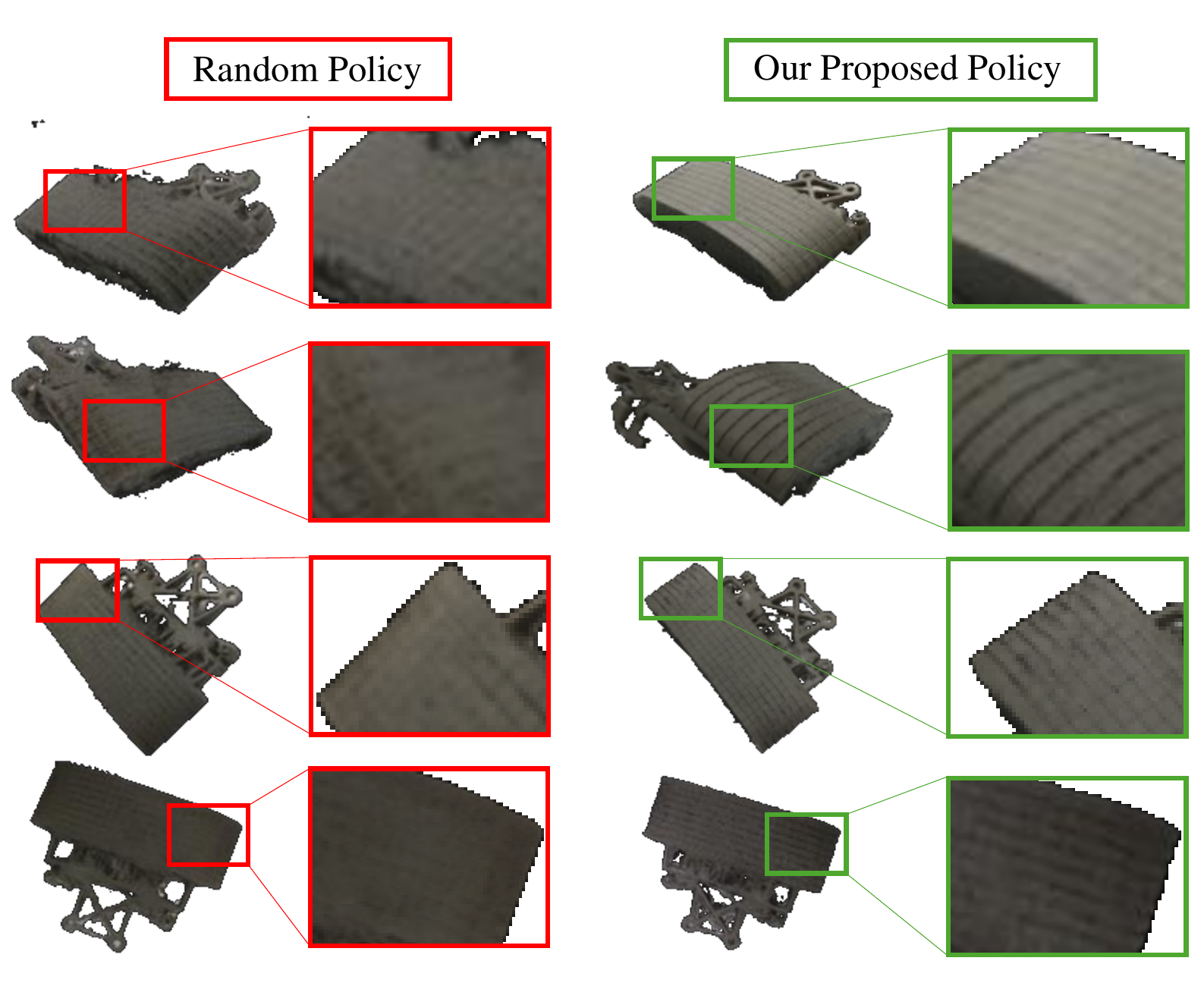}
    \caption{Comparison of local rendering quality: Proposed algorithm vs. random viewpoint selection.}
    \label{fig: Comapre_rendering}
\end{figure}
To further validate \gls{rlhf}’s effectiveness, we compare the local rendering quality between our approach and random viewpoint selection strategies. Fig.~\ref{fig: Comapre_rendering} highlights instances where our proposed algorithm viewpoint selection achieves sharper details in occluded or complex regions, whereas random selection often fails to capture fine-grained details. These results confirm that our proposed algorithm not only improves overall scene representation but also enhances local rendering accuracy, making it particularly suitable for applications requiring precise reconstruction of critical areas in nuclear decommissioning and remote inspection tasks.

\section{Conclusions}
\label{sec:conclusions}
This paper presented an \gls{rlhf}-based framework for active 3D scene representation, aligning robotic viewpoint planning with human preferences. Our approach bridged the gap between geometry-driven methods and user-specific needs, optimizing viewpoint selection in safety-critical environments. Experiments in a nuclear decommissioning scenario demonstrated significant improvements in scene representation quality, viewpoint efficiency, and alignment with operator objectives. Future work included scalable human feedback integration, leveraging pretrained language models for high-level reasoning, and improving uncertainty modelling for constrained sensing conditions. This study highlighted the potential of human-in-the-loop optimization in robotics, paving the way for more intuitive and efficient autonomous exploration.

\section{Acknowledgment}
The authors would like to express their sincere gratitude to the  Robotics and AI Collaboration (RAICo), UK Atomic Energy Authority (UKAEA), and Sellafield Ltd for their invaluable support throughout this research. In particular, we extend our appreciation to Remote Applications in Challenging Environments (RACE) and the JET Decommissioning and Repurposing (JDR) team in UKAEA for their support and guidance. We are also especially grateful to Salvador Pacheco-Gutierrez, Robert Howell, and Craig West for their insightful discussions and constructive feedback.

\normalem
\bibliography{IEEEabrv,bib}
\bibliographystyle{IEEEtran}

\end{document}